\documentclass[10pt,twocolumn,letterpaper]{article}

\usepackage{iccv}
\makeatletter
\@namedef{ver@everyshi.sty}{}
\makeatother
\usepackage{times}
\usepackage{epsfig}
\usepackage{graphicx}
\usepackage{amsmath}
\usepackage{amssymb}

\usepackage{multirow}
\usepackage{tikz}
\usepackage{booktabs}
\usepackage{float}
\usepackage{subfig}
\usepackage{makecell}
\usepackage{authblk}

\usepackage[pagebackref=true,breaklinks=true,letterpaper=true,colorlinks,bookmarks=false]{hyperref}

\iccvfinalcopy 

\def\iccvPaperID{****} 

\ificcvfinal\fi

\newcommand\modelname{{KFV}\xspace}
\DeclareMathOperator*{\argmax}{arg\,max}

\begin{document}

\title{Knowledge-augmented Few-shot Visual Relation Detection}

\author[1]{Tianyu Yu}
\author[1]{Yangning Li}
\author[2]{Jiaoyan Chen}
\author[1]{Yinghui Li}
\author[1]{Hai-Tao Zheng$^\dag$}
\author[3]{Xi Chen$^\dag$}
\author[3]{Qingbin Liu}
\author[3]{Wenqiang Liu}
\author[3]{Dongxiao Huang}
\author[3]{Bei Wu}
\author[3]{Yexin Wang}
\affil[1]{Shenzhen International Graduate School, Tsinghua University, Shenzhen, China}
\affil[2]{Department of Computer Science, The University of Manchester, Mancester, UK}
\affil[3]{Tencent, Shenzhen, China}
\affil[ ]{\texttt{yiranytianyu@gmail.com}}

\maketitle
\ificcvfinal\thispagestyle{empty}\fi

\begin{abstract}
    Visual Relation Detection (VRD) aims to detect relationships between objects for image understanding.
    Most existing VRD methods rely on thousands of training samples of each relationship to achieve satisfactory performance.
    Some recent papers tackle this problem by few-shot learning with elaborately designed pipelines and pre-trained word vectors. However, the performance of existing few-shot VRD models is severely hampered by the poor generalization capability, as they struggle to handle the vast semantic diversity of visual relationships.
    Nonetheless, humans have the ability to learn new relationships with just few examples based on their knowledge.
    Inspired by this, we devise a knowledge-augmented, few-shot VRD framework leveraging both textual knowledge and visual relation knowledge to improve the generalization ability of few-shot VRD. The textual knowledge and visual relation knowledge are acquired from a pre-trained language model and an automatically constructed visual relation knowledge graph, respectively.
    We extensively validate the effectiveness of our framework.
    Experiments conducted on three benchmarks from the commonly used Visual Genome dataset show that our performance surpasses existing state-of-the-art models with a large improvement.
\end{abstract}
{\let\thefootnote\relax\footnotetext{$\dag$ Corresponding authors: zheng.haitao@sz.tsinghua.edu.cn, jasonxchen@tencent.com}}


\section{Introduction}

\begin{figure}[t]
    \centering
    \includegraphics[width=\columnwidth]{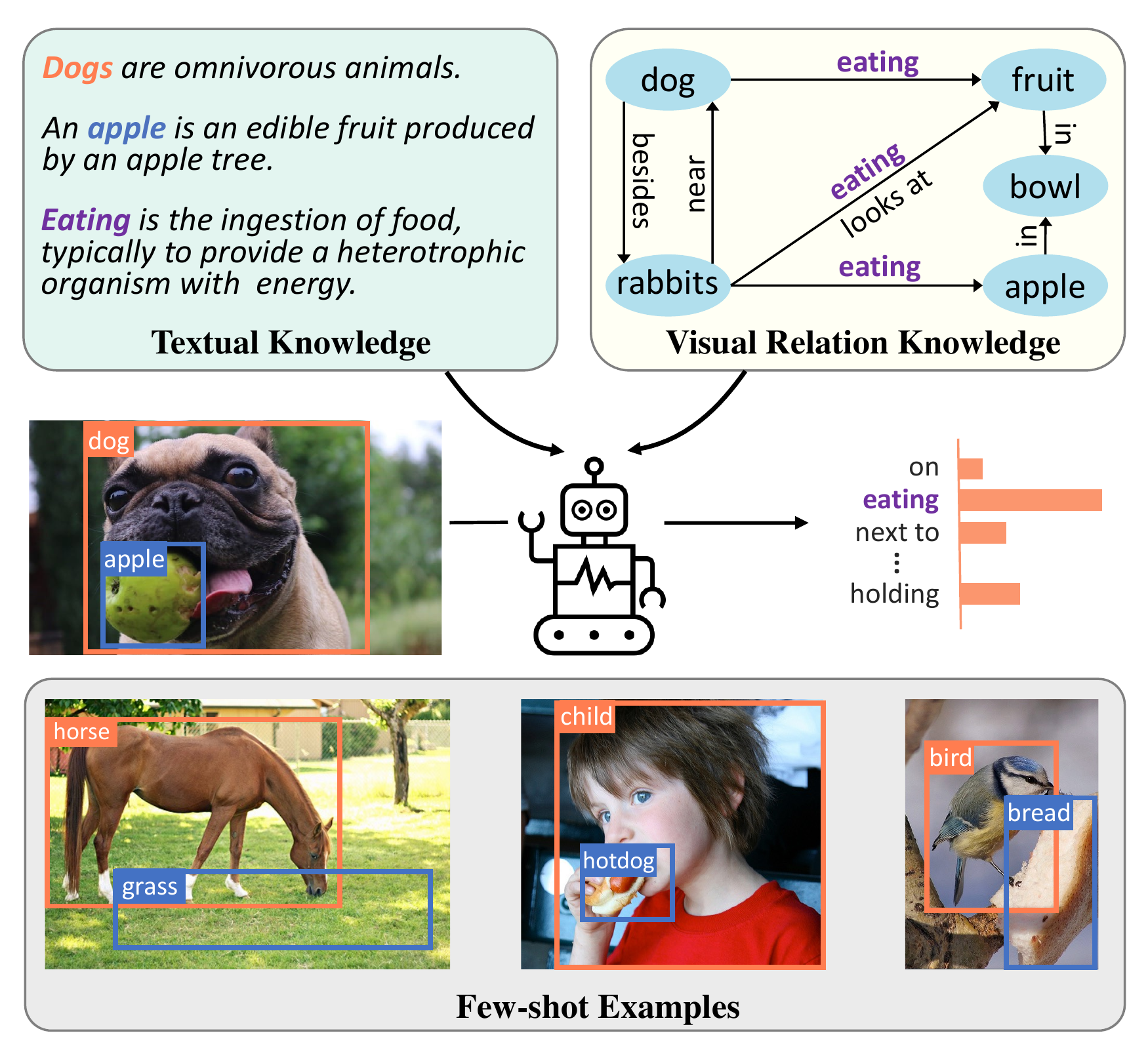}
    \caption{Leveraging external knowledge to augment few-shot VRD. Due to the large semantic diversity of \texttt{eating}, a model trained with the few-shot examples, such as \texttt{horse-eating-grass} or \texttt{child-eating-hotdog} cannot be easily generalized to new examples, such as \texttt{dog-eating-apple}. While external knowledge can be leveraged to improve the model's generalization ability. For instance, textual knowledge from large textual corpora can describe meanings of \texttt{dog}, \texttt{apple}, and \texttt{eating}.
    Visual relation knowledge learnt from the visual relation knowledge graph can show common relationships between objects, such as the fact that \texttt{eating} is a common relationship between home pets (\eg, \texttt{dog}, \texttt{rabbits}) and fruits (\eg, \texttt{apple}).}
    \label{fig:demo}
\end{figure}

Visual Relation Detection (VRD) targets at the detection of relations, \ie \texttt{Subject}-\texttt{Predicate}-\texttt{Object} triplets, which capture a wide variety of interactions (relationships) between objects in images. For example, the input image in Figure \ref{fig:demo} shows the \texttt{eating} relationship between the \texttt{dog} and the \texttt{apple}. The detected relations provide a deep and comprehensive understanding of the semantic content of images and have facilitated state-of-the-art models in numerous applications in computer vision such as visual question answering\cite{hudson2019learning, teney2017graph}, image retrieval \cite{johnson2015image}, and image captioning \cite{yang2019auto, li2019know}.

Most existing VRD methods \cite{lyu2022fine, li2022sgtr, dong2022stacked} rely on a large amount of training data for each relationship to achieve satisfactory performance. However, the distribution of relationships in real-world images is extremely long-tail since many relationships are inherently uncommon. Specifically, 92.3\% of relationships in the Visual Genome dataset \cite{krishnavisualgenome} have only ten or fewer samples (\eg, \texttt{packing}, \texttt{spouting}), while \texttt{wearing} has about 12,400 times more samples. As a result, these methods can only handle those frequent relationships which have many labeled samples.
In order to enable VRD with limited samples, some recent works try to solve this task in the few-shot setting with either visual relation detection pre-training on frequent relationships \cite{dornadula2019visual} or elaborately designed message passing strategies \cite{wang2020one} or lexical knowledge of object class tags \cite{li2022zero}.
However, these approaches cannot achieve satisfactory generalization ability given the large semantic diversity of a relationship, which is caused by the compositional nature.
To be specific, given a certain relationship, the classes of \texttt{Subject} and \texttt{Object} can be diverse, resulting huge semantic difference between relation triplets of this relationship.
For example, the difference between an image showing \texttt{dog-eating-apple} and an image showing \texttt{bird-eating-bread} can be huge since the outlook of \texttt{dog} and \texttt{bird} varies dramatically and their eating actions are also different. Inspired by the fact that human can distinguish new visual relationships efficiently with only a small number of samples based on human knowledge \cite{grishman2005nyu}, we introduce a knowledge augmented few-shot VRD framework demonstrated in Figure \ref{fig:demo}. Our framework leverages both textual knowledge and visual relation knowledge, and can largely improve the generalization ability.

\paragraph{Textual knowledge.}
Text conveys the knowledge about most entities, relationships, which can be utilized to mitigate the semantic diversity problem and consequently improve the generalization ability of few-shot VRD models.  With self-supervised training on large textual corpora, pre-trained language models (PLMs) are regarded as the most common source of textual knowledge and are shown to have knowledge about entities \cite{broscheit2020investigating} and their relations \cite{petroni2019language}. Rather than representing each relationship statically with few-shot training examples \cite{wang2020one} or word vectors \cite{li2022zero}, we propose to generate the knowledge-enhanced representation of relationships dynamically with a pre-trained language model, based on the classes of \texttt{Subject} and \texttt{Object} and a simple prompt template.
As a result, the semantic diversity of each relationship can be largely covered with the help of the embedded knowledge in the pre-trained language model.

\paragraph{Visual Relation Knowledge.} Though convey a large amount of useful information, knowledge from text suffer from the reporting bias problem \cite{gordon2013reporting} and misses common relation facts such as \texttt{letter-painted on-sign} and \texttt{sign-hanging from-pole}. In particular, the well-known text-based knowledge graph ConceptNet \cite{speer2017conceptnet} covers only 17\% of relation facts in visual relation learning datasets \cite{yao2023clever}. However, visual relation knowledge which explicitly represents the common relationships between objects can be leveraged to bridge the information of entities and relationships from the textual knowledge. To acquire such knowledge, we propose to automatically construct a visual relation knowledge graph from tremendous open domain image captions \cite{lin2014microsoft}. In detail, we adopt a rule-based parser \cite{wu2019parser} to extract relation triplets from captions and then resort to a distributed representation of visual relation knowledge graph with a novel knowledge encoding module.


Our contributions can be summarized as three-fold:
\begin{itemize}
    \item  To the best of our knowledge, this is the first work to combine textual knowledge and visual relation knowledge in VRD. We devise a \underline{K}nowledge-augmented \underline{F}ew-shot \underline{V}RD (\modelname) framework, which incorporates both kinds of knowledge to improve the generalization ability of few-shot VRD.
    \item We present a novel prompt-based method to acquire textual knowledge from pre-trained language models and an efficient way to learn distributed visual relation knowledge graph representation from image captions to augment few-shot VRD.
    \item We conduct comprehensive experiments which demonstrate the effectiveness of the proposed framework, which outperforms existing state-of-the-art models by a large margin on three benchmarks from the Visual Genome dataset.
\end{itemize}

\begin{figure*}
    \centering
    \includegraphics[width=\textwidth]{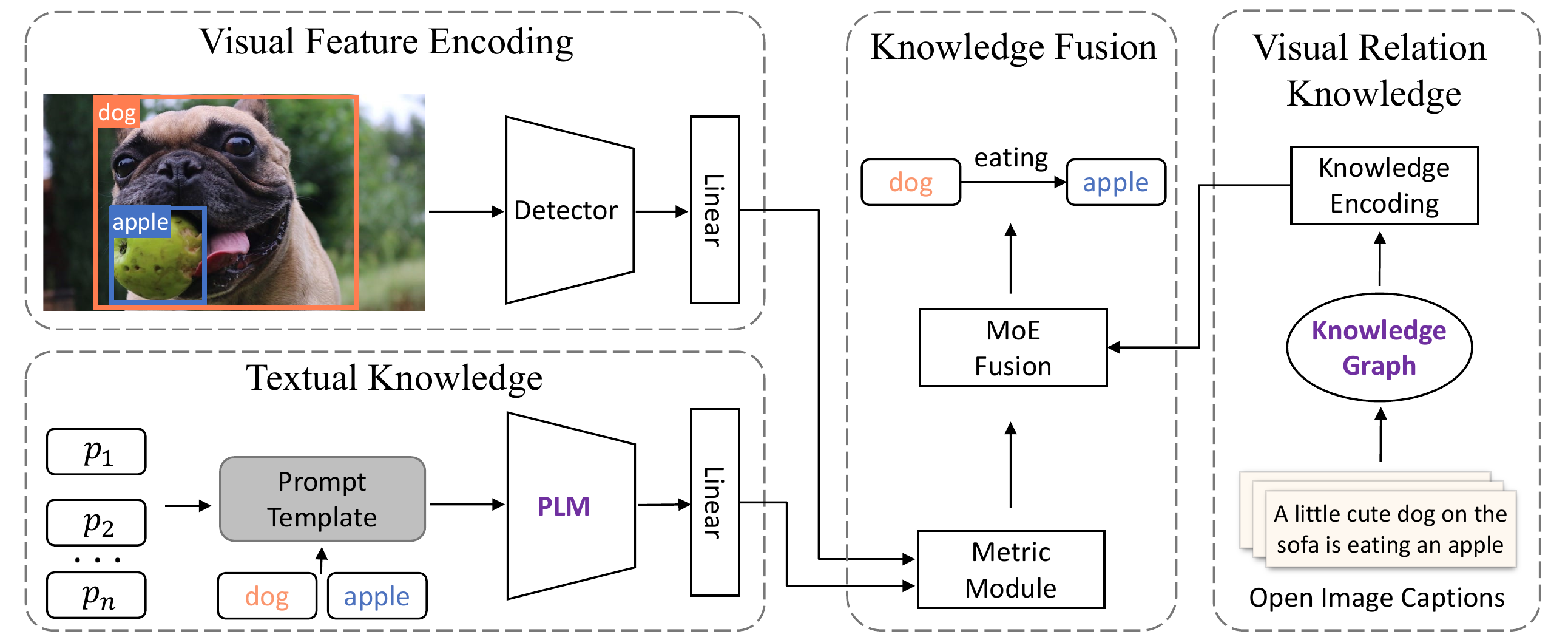}
    \caption{The architecture of our proposed \modelname framework. The framework first acquires textual knowledge and visual relation knowledge related to the input sample and then fuses them to generate the final prediction.
    }
    \label{fig:architecture}
\end{figure*}

\section{Problem Definition}

In this section, we formally define few-shot VRD and introduce related annotations to be used in the following sections.
Given an image $I$, the corresponding objects $\mathcal{O} = \{o_1, o_2, \cdots, o_n\}$ in this image, and a set of candidate relationships $\mathcal{P} = \{p_1, p_2, \cdots, p_m\}$, VRD is to develop a model to detect all possible relationships between each object pair $(o_i, o_j)$. Note each object $o_i = (b_i, c_i)$ contains a bounding box $b_i$ and a class tag $c_i$ from $\mathcal{C} = \{c_1, c_2, \cdots, c_N\}$, and a special \texttt{no-relation} relationship is also included in $\mathcal{P}$, indicating there is no relationship between an object pair.
In few-shot VRD, only a very small number of examples are given for each relationship $p_k$ for developing the VRD model.

\section{Methodology}

The overall architecture of our proposed framework is shown in Figure \ref{fig:architecture}. We start by introducing the visual feature encoding module in Section \ref{sec:method_visual_feat_enc}. Then in Section \ref{sec:method_textual-knowledge}, we introduce how to acquire textual knowledge from a pre-trained language model. In Section \ref{sec:method_visual-relation-knowledge}, we show how to utilize the visual relation knowledge from image captions. Finally, we introduce how to fuse knowledge from different sources in Section \ref{sec:method_knowldge_fusion} and explain the training and inference detail in Section \ref{sec:method_train_and_inference}.

\subsection{Visual Feature Encoding}
\label{sec:method_visual_feat_enc}

Visual features including the appearance information and spatial information of each object are essential for VRD. We first pass the image $I$ and boxes $b_1, b_2, \cdots, b_n$ to a pre-trained object detector used in \cite{zhang2021vinvl} to obtain the raw visual features $\bold{o}^{r}_1, \bold{o}^{r}_2, \cdots, \bold{o}^{r}_n$. Then a linear layer is used to encode the raw features to get the final representation of each object pair which contains the appearance and the spatial information of both objects:
\begin{equation}
    \bold{v}_{ij} = W_v \cdot \operatorname{concat}(\bold{o}_i, \bold{o}_j) + b_v ,
\end{equation}
where $W_v \in \mathbb{R}^{2D_v \times H}$ and $b_v \in \mathbb{R}^{H}$ are trainable parameters, $D_v$ and $H$ are the sizes of the raw features and the output visual features, respectively.

\subsection{Textual Knowledge}
\label{sec:method_textual-knowledge}

\begin{table*}[t]
    \centering
    \begin{tabular}{|c|c|c|c|}
    \hline
    \textbf{Input} & \textbf{Template} & \textbf{Example} & \textbf{Result Prompt}  \\
    \hline
          \multirow{3}{*}{\makecell{(\texttt{dog}, \texttt{apple}) \\ \texttt{eating}}}
          & Cloze-style\cite{schick2020exploiting} & \makecell{The relationship between \texttt{Subject} \\ and \texttt{Object} is \texttt{Predicate}} & \makecell{The relationship between \texttt{dog} \\ and \texttt{apple} is \texttt{eating}} \\
    \cline{2-4}
          & T5-generated \cite{gao2021making} & \makecell{\texttt{Subject} and \texttt{Object} are \\ \texttt{Predicate}.} & \texttt{dog} and \texttt{apple} are \texttt{eating}. \\
    \cline{2-4}
          & Triplet (ours) & \texttt{Subject} is \texttt{Predicate} \texttt{Object}  & \texttt{dog} is \texttt{eating} \texttt{apple} \\
    \hline
    \end{tabular}
    \vspace{0.3em}
    \caption{Examples of different templates used to acquire textual knowledge from pre-trained language models.}
    \label{tab:template_demo}
\end{table*}

Textual knowledge provides general understanding of object classes, relationships, and their combinations which can be helpful to mitigate the large semantic diversity problem in VRD. Inspired by \cite{bhargava2022commonsense}, we propose to dynamically generate textual prompts to acquire such knowledge from a pre-trained language model. Examples of different templates are shown in Table \ref{tab:template_demo}.

Firstly, given the class tags $(c_i, c_j)$ of $(o_i, o_j)$ pair, and the candidate relationship $p_k$, we adopt a template to combine these parts into a textual prompt $x$.
For example, given the class tags $(\texttt{dog}, \texttt{apple} )$, the candidate relationship \texttt{eating} and a template ``\textit{the relationship between \texttt{S} and \texttt{O} is \texttt{P}}'', the prompt we get will be ``\textit{the relationship between \texttt{dog} and \texttt{apple} is \texttt{eating}}''. Secondly, we parse the prompt into a list of tokens $x_1, x_2, \cdots, x_{n_x}$ and pass the list to a pre-trained language model to obtain the knowledgeable contextualized representations of each token, denoted as $\bold{x}_1, \bold{x}_2, \cdots, \bold{x}_{n_x}$. Thirdly, we average the representation of tokens corresponding to $p_k$ to get the raw feature of $p_k$ as $\bold{p}^r_k$, which is conditioned on the generated prompt:

\begin{equation}
    \bold{p}^r_k = \frac{\sum_{x_i \in p_k}{\bold{x}_i}}{\operatorname{length}(p_k)},
\end{equation}
where $\operatorname{length}(p_k)$ is the number of tokens in $p_k$. Finally, we project $\bold{p}^r_k$ to the same dimension as $\bold{v}_{ij}$ with a linear layer to get $\bold{p}_k \in \mathbb{R}^{H}$:

\begin{equation}
    \bold{p}_k = W_p \cdot \bold{p}^r_k + b_p,
\end{equation}
where $W_p \in \mathbb{R}^{D_t}$ and $b_p \in \mathbb{R}^H$ are trainable parameters, $D_t$ is the dimension of the representation generated by the pre-trained language model.

\subsection{Visual Relation Knowledge}
\label{sec:method_visual-relation-knowledge}

Although the textual knowledge from pre-trained language models is helpful for VRD, it is known to be biased \cite{gordon2013reporting}. To mitigate these shortcomings, we propose to exploit image captions to collect visual relation knowledge which is an valuable supplementary knowledge resource and is relatively objective, since captions capture the global context of images and can serve as a powerful knowledge resource \cite{yao2021visual}.  Specifically, we extract visual relation triplets from captions and further construct a visual relation knowledge graph. An example of visual relation knowledge graph is demonstrated in Figure \ref{fig:demo}.

We utilize a widely used rule-based triplet parser \cite{wu2019parser} to extract triplets from image captions. For instance, we can get \texttt{dog-is\_eating-apple} and \texttt{dog-on-sofa} from the caption ``\textit{A little cute dog on the sofa is eating an apple}''. These triplets directly reflect the interactions among objects in the corresponding image. We then merge duplicate triplets to construct a visual relation knowledge graph denoted as $G=(\mathcal{V}, \mathcal{E}, \mathcal{R})$ with nodes (entities) $v_i \in \mathcal{V}$ and distinct triplets as edges $(v_i, r, v_j) \in \mathcal{E}$, where $r\in\mathcal{R}$ is a relationship. However, the application of $G$ still faces two problems. First, the representation of entity and relationship names can be diverse due the flexibility of natural language (\eg, \texttt{eat}, \texttt{eats} and \texttt{is\_eating}), which makes it difficult to align them to classes and predicates. Second, though parsed large amount of captions, the symbolic graph is still sparse compared to the large space of possible relation triplets.  To solve these problems and improve the generalization ability, we leverage a pre-trained BERT model \cite{devlin2018bert} to convert the graph into distributed representations. A pre-trained BERT model can naturally project semantically similar terms to be close to each other in the latent space and its compositional generalization ability \cite{herzig2021unlocking} can be leveraged to mitigate the sparsity of symbolic knowledge.

\begin{table*}[t]
\centering
\begin{tabular}{llcc ccc ccc cc cc}
\toprule
\multirow{2}{*}{\textbf{Shot}} & \multirow{2}{*}{\textbf{Model}} &\multirow{2}{*}{\textbf{OD.}} &\multirow{2}{*}{\textbf{VL.}} &\multicolumn{3}{c}{\textbf{Recall@k}} & \multicolumn{3}{c}{\textbf{Mean Recall@k}} & \multicolumn{2}{c}{\textbf{Seen Recall}} & \multicolumn{2}{c}{\textbf{Unseen Recall}} \\

\cmidrule(lr){5-7} \cmidrule(lr){8-10} \cmidrule(lr){11-12} \cmidrule(lr){13-14}

\ \ \ \ & & & & k=20  & k=50 & k=100 & k=20 & k=50 & k=100 & P & T & P & T \\
\midrule

\multicolumn{1}{c}{-} & \multicolumn{1}{l}{Random}   & & & \hspace{1.7mm}0.6 & \hspace{1.7mm}1.1 & \hspace{1.7mm}1.4 & \hspace{1.7mm}0.3 & \hspace{1.7mm}0.7 & \hspace{1.7mm}1.0 & \hspace{1.7mm}1.0 & \hspace{1.7mm}1.1 & \hspace{1.7mm}1.1 & \hspace{1.7mm}1.1  \\

\midrule

\multicolumn{1}{c}{\multirow{5}{*}{5}} & \multicolumn{1}{l}{MotifNet\cite{zellers2018motif}}   & \checkmark & & \hspace{1.7mm}3.5 & \hspace{1.7mm}5.0 & \hspace{1.7mm}6.1 & \hspace{1.7mm}5.1 & \hspace{1.7mm}7.0 & \hspace{1.7mm}8.3 & \hspace{1.7mm}9.6 & 21.2 & \hspace{1.7mm}4.9 & \hspace{1.7mm}5.1  \\

 &  \multicolumn{1}{l}{IMP\cite{xu2017IMP}}     & \checkmark & & \hspace{1.7mm}5.1 & \hspace{1.7mm}7.1 & \hspace{1.7mm}8.4 & 11.5 & 14.9 & 17.1 & 13.2 & 34.0 & \hspace{1.7mm}6.7 & \hspace{1.7mm}6.9  \\

& \multicolumn{1}{l}{CLIP\cite{radford2021CLIP}}    & & \checkmark &  \hspace{1.7mm}8.6   & 11.6 &  13.4   &  12.4   &  16.5   & 19.6    & 22.2    & \textbf{63.7}    & 11.2    & 11.1     \\

& \multicolumn{1}{l}{VinVL\cite{zhang2021vinvl}}    & \checkmark & \checkmark &  \hspace{1.7mm}5.9   & \hspace{1.7mm}8.1 &  \hspace{1.7mm}9.5   &  14.2   &  17.4   & 19.5    & 14.3    & 45.2    & \hspace{1.7mm}8.0    & \hspace{1.7mm}7.6     \\

& \multicolumn{1}{l}{\modelname (ours)}       & \checkmark & & \textbf{14.6} & \textbf{20.2} & \textbf{23.3} & \textbf{22.0} & \textbf{28.2} & \textbf{31.7} & \textbf{27.5} & 61.3 & \textbf{21.9} & \textbf{21.3} \\

\midrule

\multicolumn{1}{c}{\multirow{3}{*}{10}} & \multicolumn{1}{l}{CLIP\cite{radford2021CLIP}}    & & \checkmark &  10.5   & 13.6 &  15.4   &  15.3   &  19.8   & 22.4    & 23.5    & 63.7    & 10.8    & \hspace{1.7mm}9.5     \\

& \multicolumn{1}{l}{VinVL\cite{zhang2021vinvl}}    & \checkmark & \checkmark &  \hspace{1.7mm}7.2   & \hspace{1.7mm}9.6 &  11.0   &  18.6   &  22.7   & 25.4    & 16.0    & 44.3    & \hspace{1.7mm}7.1    & \hspace{1.7mm}6.2    \\

& \multicolumn{1}{l}{\modelname (ours)}       & \checkmark & &\textbf{14.9}&\textbf{19.9}&\textbf{22.6} &\textbf{22.9}&\textbf{28.5}&\textbf{31.6} &\textbf{26.6}&\textbf{65.4}&\textbf{20.1}&\textbf{17.3} \\

\midrule

\multicolumn{1}{c}{\multirow{3}{*}{15}} & \multicolumn{1}{l}{CLIP\cite{radford2021CLIP}}    & & \checkmark &  10.0   & 13.1 &  14.9   &  16.9   &  21.3   & 24.5    & 19.8    & 52.7    & 10.9    & \hspace{1.7mm}9.4     \\

& \multicolumn{1}{l}{VinVL\cite{zhang2021vinvl}}    & \checkmark & \checkmark &  \hspace{1.7mm}9.5   & 12.3 &  13.9   &  21.2   &  25.8   & 29.0    & 17.7    & 47.9    & \hspace{1.7mm}9.3    & \hspace{1.7mm}8.0    \\

& \multicolumn{1}{l}{\modelname (ours)}       & \checkmark & &\textbf{18.0}&\textbf{24.2}&\textbf{27.3} &\textbf{23.4}&\textbf{31.1}&\textbf{33.8} &\textbf{29.8}&\textbf{65.3}&\textbf{25.0}&\textbf{21.8} \\

\bottomrule
\end{tabular}
\vspace{0.3em}
\caption{Performance (\%) on the 50-way benchmark. OD.: object detector. VL.: vision-language pre-training. P: \texttt{Subject}-\texttt{Object} pair, T: \texttt{Subject}-\texttt{Predicate}-\texttt{Object} triplet. We report mean performance over 3 random trials.}
\label{table:50_way_main}
\end{table*}

\paragraph{Knowledge Encoding.} Specifically, we fine-tune the pre-trained BERT model to reconstruct triplets in the visual knowledge graph. Given an edge $(v_i, r, v_j)$ such as $(\texttt{dog}, \texttt{is eating}, \texttt{apple})$ from the graph, we first convert it into a natural language sentence ``dog is eating apple''. Next, replace the $r$ with a special \texttt{[MASK]} token and input the masked sentence to the pre-trained BERT. The model is trained to reconstruct the corresponding relationship $r$ from the candidate relationship set $\mathcal{R}$ with the output feature of \texttt{[MASK]}, denoted as $\bold{m}$. For relationships with multiple tokens, we empirically find that taking the average of tokens yields a good representation. With the fine-tuned BERT model as the visual relation knowledge encoder, we can dynamically acquire the visual relation knowledge of input object class pairs. Specifically, given input class tags $(c_i, c_j)$, we convert it into a natural language sentence ``$c_i \text{\texttt{[MASK]}} c_j$'', where the \texttt{[MASK]} represent the possible relationships, and input it to the encoder to get the output feature of the \texttt{[MASK]} token as $\bold{m}$. We then use $\bold{m}$ to generate a prior score for each candidate predicate $p_k$ demonstrating how likely the triplet \texttt{$c_i$-$p_k$-$c_j$} is in the visual relation knowledge graph :
\begin{equation}
\label{equ:score_visual}
    s^{v}_k = \bold{m} \cdot \frac{\sum_{x_i\in p_k}{\operatorname{Embed_{\text{KE}}}(x_i)}}{\operatorname{length}(p_k)},
\end{equation}
where $x_i$ is a token from $p_k$, $\operatorname{Embed_{\text{KE}}}$ maps tokens to corresponding output embedding.

\subsection{Knowledge Fusion}
\label{sec:method_knowldge_fusion}

In this section, we will introduce how to fuse textual knowledge and visual relation knowledge to generate the prediction. Firstly, we adopt a metric module to calculate the distance between candidate relationship $p_k$ and the input object pair $(o_i, o_j)$ representations:
\begin{equation}
\label{equ:score_text}
    s^t_k = \text{distance}(\operatorname{concat}(\bold{o}_i, \bold{o}_j), \bold{p}_k),
\end{equation}
where the distance function calculate the cosine distance between two vectors. Then we devise a novel Mixture-of-Experts (MoE) \cite{masoudnia2014mixture}  module to combine $\bold{s}^v$ and $\bold{s}^t$ to generate the final probability distribution:

\begin{equation}
    \bold{s} = \operatorname{softmax}(W_f \cdot \operatorname{concat}(\bold{s}^v, \bold{s}^t) + b_f) ,
\end{equation}
where $W_f \in \mathbb{R}^{n \times 2n}$ and $b_f \in \mathbb{R}^{n}$ are trainable parameters. $\bold{s}^v$ and $\bold{s}^t$ are calculated with Equation \ref{equ:score_visual} and Equation\ref{equ:score_text}.

\subsection{Training and Inference}
\label{sec:method_train_and_inference}

The final probability distribution $\bold{s}$ is optimized with a cross-entropy loss as follows:

\begin{equation}
    \mathcal{L} = -\log{(\mathbf{s}_*)}.
\end{equation}

During inference, we predict the visual relation (including \texttt{no-relation}) for each object pair $(o_i, o_j)$:

\begin{equation}
    p' = \argmax_{p_k \in \mathcal{P}}(\bold{s}_k).
\end{equation}

\begin{table*}[t]
\centering
\resizebox{\linewidth}{!}{
\begin{tabular}{llccc ccc ccc cc cc}
\toprule
\multirow{2}{*}{\textbf{Shot}} & \multirow{2}{*}{\textbf{Model}} &\multirow{2}{*}{\textbf{OD.}} &
\multirow{2}{*}{\textbf{VR.}} &\multirow{2}{*}{\textbf{VL.}} &\multicolumn{3}{c}{\textbf{Recall@k}} & \multicolumn{3}{c}{\textbf{Mean Recall@k}} & \multicolumn{2}{c}{\textbf{Seen Recall}} & \multicolumn{2}{c}{\textbf{Unseen Recall}} \\

\cmidrule(lr){6-8} \cmidrule(lr){9-11} \cmidrule(lr){12-13} \cmidrule(lr){14-15}

\ \ \ \ & & & & & k=20  & k=50 & k=100 & k=20 & k=50 & k=100 & P & T & P & T \\
\midrule

\multicolumn{1}{c}{-} & \multicolumn{1}{l}{Random}   & & & & \hspace{1.7mm}0.9 & \hspace{1.7mm}1.8 & \hspace{1.7mm}2.6 & \hspace{1.7mm}0.9 & \hspace{1.7mm}1.6 & \hspace{1.7mm}2.4 & \hspace{1.7mm}2.5 & \hspace{1.7mm}2.5 & \hspace{1.7mm}2.0 & \hspace{1.7mm}1.9  \\

\midrule

\multicolumn{1}{c}{\multirow{7}{*}{5}} & \multicolumn{1}{l}{LPaF* \cite{dornadula2019visual}}    & \checkmark & \checkmark & & -   & 20.9 &  -   &  -   &  -   & -    & -    & -    & -    & -     \\

& \multicolumn{1}{l}{LKMN* \cite{li2022zero}}    & \checkmark & \checkmark & &  -   & 22.0 &  -   &  -   &  -   & -    & -    & -    & -    & -     \\

 & \multicolumn{1}{l}{MotifNet\cite{zellers2018motif}}   & \checkmark & & & 13.2 & 17.7 & 20.0 & 13.2 & 17.0 & 19.2 & 34.6 & 45.8 & 15.0 & 14.6  \\

 & \multicolumn{1}{l}{IMP\cite{xu2017IMP}}     & \checkmark & & & 22.5 & 27.3 & 29.7 & 23.3 & 27.6 & 30.0 & 47.0 & 63.2 & 24.0 & 22.9  \\

& \multicolumn{1}{l}{CLIP\cite{radford2021CLIP}}    & & & \checkmark &  25.2   & 30.8 &  34.0   &  24.9   &  30.8   & 34.3    & \textbf{63.3}    & \textbf{88.5}    & 24.1    & 22.4     \\

& \multicolumn{1}{l}{VinVL\cite{zhang2021vinvl}}    & \checkmark & & \checkmark &  24.8   & 30.2 &  33.1   &  26.4   &  31.4   & 34.1    & 59.8    & 82.9    & 24.0    & 22.4     \\

& \multicolumn{1}{l}{\modelname (ours)}       & \checkmark & & & \textbf{33.2} & \textbf{40.3} & \textbf{43.6} & \textbf{33.1} & \textbf{40.2} & \textbf{43.7} & 61.4 & 83.8 & \textbf{37.2} & \textbf{34.7} \\

\midrule

\multicolumn{1}{c}{\multirow{3}{*}{10}} & \multicolumn{1}{l}{CLIP\cite{radford2021CLIP}}    & & & \checkmark &29.9&36.5&39.9 &28.0&35.4&39.1 &62.4&\textbf{86.5}&27.3&24.2     \\

& \multicolumn{1}{l}{VinVL\cite{zhang2021vinvl}}    & \checkmark & & \checkmark &29.2&35.1&38.4 &31.0&36.9&40.6 &55.4&74.1&27.9&25.5     \\

& \multicolumn{1}{l}{\modelname (ours)}       & \checkmark & & &\textbf{37.8}&\textbf{44.3}&\textbf{48.3} &\textbf{38.5}&\textbf{45.5}&\textbf{49.5} &\textbf{63.1}&83.2&\textbf{39.8}&\textbf{36.4} \\

\midrule

\multicolumn{1}{c}{\multirow{3}{*}{15}} & \multicolumn{1}{l}{CLIP\cite{radford2021CLIP}}    & & & \checkmark &32.8&38.7&41.9 &32.3&39.4&42.5 &58.6&\textbf{78.9}&30.8&26.8     \\

& \multicolumn{1}{l}{VinVL\cite{zhang2021vinvl}}    & \checkmark & & \checkmark &33.0&39.9&43.3 &33.8&41.2&44.5 &55.7&73.9&32.6&29.3    \\

& \multicolumn{1}{l}{\modelname (ours)}       & \checkmark & & &\textbf{40.1}&\textbf{47.1}&\textbf{50.3} &\textbf{39.3}&\textbf{47.5}&\textbf{50.9} &\textbf{60.8}&78.2&\textbf{42.2}&\textbf{38.0} \\

\bottomrule
\end{tabular}}
\vspace{0.1mm}
\caption{Performance (\%) on the 25-way benchmark. OD.: object detector. VR.: visual relation detection pre-training, VL.: vision-language pre-training. P: \texttt{Subject}-\texttt{Object} pair, T: \texttt{Subject}-\texttt{Predicate}-\texttt{Object} triplet. We report mean performance over 3 random trials. *: Results of LPaF and LKMN are obtained from corresponding papers directly.}
\label{table:25_way_main}
\end{table*}

\section{Experiments and Evaluation}


\begin{table*}[t]
\centering
\resizebox{\linewidth}{!}{
\begin{tabular}{clccc ccc ccc cc cc}
\toprule
\multirow{2}{*}{\textbf{Shot}} & \multirow{2}{*}{\textbf{Model}} &\multirow{2}{*}{\textbf{OD.}} &\multirow{2}{*}{\textbf{VL.}} &\multirow{2}{*}{\textbf{WS.}}&\multicolumn{3}{c}{\textbf{Recall@k}} & \multicolumn{3}{c}{\textbf{Mean Recall@k}} & \multicolumn{2}{c}{\textbf{Seen Recall}} & \multicolumn{2}{c}{\textbf{Unseen Recall}} \\

\cmidrule(lr){6-8} \cmidrule(lr){9-11} \cmidrule(lr){12-13} \cmidrule(lr){14-15}

\ \ \ \ & & & & & k=20  & k=50 & k=100 & k=20 & k=50 & k=100 & P & T & P & T \\
\midrule

\multicolumn{1}{c}{\multirow{2}{*}{-}} & \multicolumn{1}{l}{Random}   & & & & \hspace{1.7mm}0.8 & \hspace{1.7mm}1.5 & \hspace{1.7mm}2.5 & \hspace{1.7mm}0.8 & \hspace{1.7mm}1.3 & \hspace{1.7mm}2.2 & \hspace{1.7mm}2.4 & \hspace{1.7mm}2.2 & \hspace{1.7mm}1.9 & \hspace{1.7mm}2.1  \\

& \multicolumn{1}{l}{VisualDS* \cite{yao2021visual}}    & \checkmark & \checkmark & \checkmark & -   & 53.4 &  56.5   &  -   &  37.7   & 41.2    & -    & -    & -    & -     \\

\midrule

\multicolumn{1}{c}{\multirow{5}{*}{5}} & \multicolumn{1}{l}{MotifNet\cite{zellers2018motif}}   & \checkmark & & & 13.9 & 17.2 & 19.2 & 14.8 & 19.1 & 21.6 & 27.8 & 35.7 & 15.5 & 15.6  \\

 & \multicolumn{1}{l}{IMP\cite{xu2017IMP}}     & \checkmark & & & 29.5 & 35.6 & 38.8 & 31.9 & 38.9 & 42.8 & 48.8 & 66.8 & 32.7 & 31.6  \\

& \multicolumn{1}{l}{CLIP\cite{radford2021CLIP}}    & & \checkmark & & 30.9   & 37.6 &  41.5   &  30.8   &  41.6   & 46.9    & 54.8    & \textbf{91.4}    & 34.2    & 30.5     \\

& \multicolumn{1}{l}{VinVL\cite{zhang2021vinvl}}    & \checkmark & \checkmark &  & 32.0   & 37.9 &  40.9   &  34.7   &  41.8   & 45.5    & 54.2    & 76.6    & 33.6    & 32.3     \\

& \multicolumn{1}{l}{\modelname (ours)}       & \checkmark & & & \textbf{42.0} & \textbf{49.4} & \textbf{53.2} & \textbf{37.7} & \textbf{45.0} & \textbf{49.4} & \textbf{61.4} & 84.2 & \textbf{46.6} & \textbf{44.5} \\

\midrule

\multicolumn{1}{c}{\multirow{4}{*}{10}} & \multicolumn{1}{l}{Limited Labels* \cite{chen2019scene}}    &\checkmark &  & \checkmark & -   & 49.7 &  50.7   &  -     & 37.4    & 38.9 &  -     & -    & -    & -     \\

 & \multicolumn{1}{l}{CLIP\cite{radford2021CLIP}}    & & \checkmark & & 35.3   & 42.7 &  46.2   &  34.4   &  44.8   & 51.3    & \textbf{62.2}    & \textbf{87.7}    & 32.9    & 29.0     \\

& \multicolumn{1}{l}{VinVL\cite{zhang2021vinvl}}    & \checkmark & \checkmark &  & 38.6   & 45.4 &  48.7   &  40.6   &  49.9   & 53.5    & 58.1    & 74.5    & 39.3    & 36.8     \\

& \multicolumn{1}{l}{\modelname (ours)}       & \checkmark & & &\textbf{43.5}&\textbf{50.7}&\textbf{54.1} &\textbf{46.3}&\textbf{54.0}&\textbf{58.1} &61.5&82.9 &\textbf{44.0}&\textbf{39.9} \\

\midrule

\multicolumn{1}{c}{\multirow{3}{*}{15}} & \multicolumn{1}{l}{CLIP\cite{radford2021CLIP}}    & & \checkmark & & 38.9   & 46.7 &  50.5   &  35.6   &  47.3   & 54.1    & \textbf{60.5}    & \textbf{85.6}    & 36.7    & 31.0     \\

& \multicolumn{1}{l}{VinVL\cite{zhang2021vinvl}}    & \checkmark & \checkmark &  & 43.6   & 50.5 &  53.5   &  42.4   &  53.2   & 56.8    & 60.0    & 80.1    & 43.3    & 38.3     \\

& \multicolumn{1}{l}{\modelname (ours)}       & \checkmark & & &\textbf{47.0}&\textbf{54.2}&\textbf{57.3} &\textbf{48.8}&\textbf{56.6}&\textbf{61.0} &\textbf{60.5}&84.4 &\textbf{48.1}&\textbf{40.7} \\

\bottomrule
\end{tabular}
}
\vspace{0.1mm}
\caption{Performance (\%) of on the 20-way benchmark. OD.: object detector. VR.: vision-language pre-training. WS.: weak supervision. P: \texttt{Subject}-\texttt{Object} pair, T: \texttt{Subject}-\texttt{Predicate}-\texttt{Object} triplet. We report mean performance over 3 random trials. *: Results of VisualDS and Limited Labels are obtained from corresponding papers directly.}
\label{table:20_way_main}
\end{table*}

\subsection{Experimental Settings}

\paragraph{Dataset.} We use the Visual Genome \cite{krishnavisualgenome} dataset for evaluation. Each image in this dataset is annotated with objects (including bounding boxes and object classes) and relations. Specifically, we use three benchmarks from this dataset: \\
\textbf{(1) 50-way}. This benchmark contains the 50 most frequent relationships and 150 object classes in Visual Genome, and is widely used in many supervised VRD works \cite{dupty2020visual, desai2021learning, plesse2018learning}.
\\
\textbf{(2) 25-way}. This benchmark is proposed in \cite{dornadula2019visual} by removing the 25 most frequent relationships from the 50-way benchmark. Many existing few-shot VRD works \cite{dornadula2019visual, li2022zero} evaluated their approach on this benchmark and use samples of removed relationships to pre-train their models.
\\
\textbf{(3) 20-way}. This benchmark is proposed by \cite{chen2019scene} which removes hypernyms and redundant synonyms in the 50-way benchmark, giving a cleaner relationship schema.

We compare our method extensively with other baselines on all three benchmarks and conduct ablation studies on the 25-way benchmark with 5 training examples.
We refer readers to the appendix for the full list of relationships used in each benchmark and other statistics.



\paragraph{Evaluation Metrics.}

We evaluate the overall VRD performance of our method using Recall@k (R@k) and mean-Recall@k (mR@k), which measure how many labeled visual relations are recalled. To support multi-dimensional and more fine-grained evaluation on the methods' generalization capability, we devise novel Seen/Unseen Recall metrics. Specifically, Seen/Unseen Recall measures the recall of visual relations that are either pair-wise or triplet-wise seen/unseen during training. Take pair-wise Unseen Recall (puR) as an example:

\begin{equation}
    \text{puR} = \frac{\sum\limits_{I\in \mathcal{I}}\sum\limits_{\texttt{s-p-o}\in \operatorname{GT}(I)}\mathbb{I}_{recall}(\texttt{s-p-o})\cdot \sigma_{pair}(\texttt{s},\texttt{o})} {\sum\limits_{I\in \mathcal{I}}\sum\limits_{\texttt{s-p-o}\in \operatorname{GT}(I)}\sigma_{pair}(\texttt{s},\texttt{o})},
\end{equation}
where $\mathbb{I}_{recall}(\texttt{s-p-o})$ is 1 if triplet $\texttt{s-p-o}$ is recalled, $\sigma_{pair}(\texttt{s},\texttt{o})$ is 1 if object class pair $(\texttt{s},\texttt{o})$ is unseen during training and 0 otherwise, $\operatorname{GT}(I)$ is labeled triplets in $I$, $\mathcal{I}$ is set of test images in the benchmark. This metric measures the ratio of testing triplets whose \texttt{Subject-Object} pair is not seen during training can be recalled. Such triplets are hard to generalize to. Similarly, we can define triplet-wise Seen Recall (tsR), pair-wise Seen Recall (psR) and triplet-wise Unseen Recall (tuR).

\paragraph{Baselines.} We compare our proposed \modelname framework with a diverse range of strong baselines including the following three categories:
\\
(1) VRD methods designed for the fully-supervised setting. \textbf{IMP}\cite{xu2017IMP} predicts relationships between object pairs with only visual features generated from an iterative message passing scheme over objects in the image.
\textbf{MotifNet}\cite{zellers2018motif} is a strong and widely adopted method that uses the local object context in an image and pre-trained word vectors to enrich the understanding of object pairs.  For fair comparison, we re-implement IMP and MotifNet with the same detector \cite{zhang2021vinvl} as \modelname.
\\
(2) Low-resource VRD methods. \textbf{LPaF} \cite{dornadula2019visual} trains a two-layer MLP to classify the input object pair. In order to improve the sample efficiency, \textbf{LKMN}\cite{li2022zero} reduces the diversity among samples with external lexical knowledge and uses pre-trained word vectors to transfer knowledge among relationships. Both LPaF and LKMN use samples of frequent relationships to pre-train the model under the fully-supervised setting (\ie VRD pre-training), thus still requiring a large amount of visual relation annotations. \textbf{Limited Labels} \cite{chen2019scene} is the state-of-the-art semi-supervised model, which learns a relational generative model using seed examples of each relationship to assign pseudo-labels to unlabeled data for training. \textbf{VisualDS} \cite{yao2021visual} is the state-of-the-art weakly supervised model, which first constructs a knowledge base from Web-scale captions to assign pseudo-labels to unlabeled data, and then leverages a pre-trained vision-language model and EM optimization to reduce the noise in pseudo-labels for training.
\\
(3) Large pre-trained vision-language  models. We compare our method with fine-tuned \textbf{CLIP} \cite{radford2021CLIP} and \textbf{VinVL}\cite{zhang2021vinvl}. CLIP was pre-trained on 400 million (image, text) pairs with an image-level contrastive objective. VinVL was pre-trained on millions of (images, text) pairs with fine-grained object-level annotations.

\subsection{Main Results}

Based on the experimental results in Table \ref{table:50_way_main} \ref{table:25_way_main}, and \ref{table:20_way_main}, we have the following observations:
(1) \modelname consistently outperforms existing state-of-the-art models across most metrics, even without any VRD pre-training, pseudo-labels, or vision-language pre-training. In particular, \modelname nearly doubles the Recall@50 score of recently release LKMN (from 22.0 to 38.4) in the 25-way-5-shot setting.
(2) Despite being trained on massive amounts of multi-modal corpus for vision-language pre-training, CLIP and VinVL do not demonstrate the same level of generalization as \modelname, especially on unseen object pairs and triplets. This highlights the necessity of explicitly incorporating knowledge from both textual domain and visual relation domain.
(3) As shown in Table \ref{table:25_way_main}, LPaF and LKMN achieve higher performance than MotifNet, which is consistent with previous studies \cite{dornadula2019visual, li2022zero}.
(4) As shown in Table \ref{table:20_way_main}, Limited Labels and Visual DS struggle to achieve satisfactory Mean Recall performance compared to other methods.  We argue that this is because the pseudo-labels can still be sparse for relationships that are inherently uncommon. To save space, we do not list the 10-shot and 15-shot experimental results for all baselines. We refer readers to the appendix for complete tables of experimental results.

\begin{table}[]
    \centering
    \begin{tabular}{l | cc}
    \toprule
        Model & psR & puR \\
    \midrule
    \modelname w/o T & 52.4 (\hspace{0.2mm}-\hspace{0.5mm}9.0) & 28.7 (-8.5) \\
    \modelname w/o V & 61.6 (+0.2) & 29.9 (-7.3)\\
    \bottomrule
    \end{tabular}
    \vspace{0.3em}
    \caption{Performance of \modelname with one knowledge source. We also show the absolute decline with respect to the results of our full model. \modelname w/o T: \modelname without textual knowledge. \modelname w/o V: \modelname without visual relation knowledge.}
    \label{tab:knowledge_ablation}
\end{table}

\begin{table}[]
    \centering
    \begin{tabular}{l | cc}
    \toprule
        Template & psR & puR \\
    \midrule
        Cloze-style \cite{schick2020exploiting} & 58.9 ($\pm$ 1.8) & 36.5 ($\pm$ 0.2) \\
        T5-generated \cite{gao2021making} & 60.5 ($\pm$ 1.2) & 35.7 ($\pm$ 0.5) \\
        Triplet (ours) & 61.4 \hspace{10mm} & 37.2 \hspace{10.5mm} \\
    \bottomrule
    \end{tabular}
    \vspace{0.3em}
    \caption{Performance of different prompt templates. We report mean (and standard deviation) performance for baselines over three examples.}
    \label{tab:prompt_ablation}
\end{table}

\subsection{Analysis of Textual Knowledge}


To further investigate the effectiveness of textual knowledge in VRD, we evaluate the performance of \modelname without using textual knowledge and list the result in Table \ref{tab:knowledge_ablation}, from which we can observe that the usage of textual knowledge can significantly improve the performance of detecting relationships of both seen and unseen object pairs. In detail, we replace the knowledgeable representation $\bold{p}_k$ with the average of pre-trained word vectors.

We also perform experiments to test different prompt templates, including manually crafted templates \cite{schick2020exploiting} following the cloze question style and automatically generated templates \cite{gao2021making} from a pre-trained T5 model. The results are presented in Table \ref{tab:prompt_ablation}. Our proposed template shows the best performance compared to baseline templates. The reason is that cloze-style prompts and T5-generated prompts consider the \texttt{Subject-Object} pair as an unit and place the \texttt{Predicate} before or after this pair which makes the direction of relation hard to be discriminated. In contrast, our proposed template places the \texttt{Predicate} between the pair, resulting a more natural prompt and explicitly represent the direction of the relation.

\begin{figure*}
    \centering
    \includegraphics[width=\linewidth]{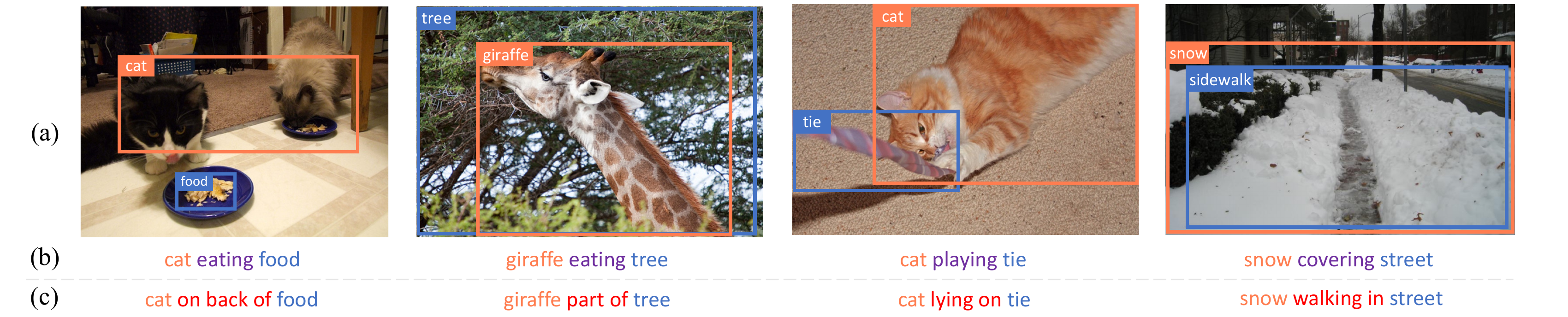}
    \caption{Examples of qualitative results. (a): input image and the object pair. (b): prediction generated by \modelname. (c): prediction generated by baseline model \cite{xu2017IMP}. Purple and red indicate correct and wrong predictions. Best viewed in color.}
    \label{fig:case}
\end{figure*}

\subsection{Analysis of Visual Relation Knowledge}

We conduct extensive experiments to analyze the impact of visual relation knowledge on our framework's performance. We first evaluate the performance of \modelname without using visual relation knowledge, specifically we directly use $\bold{s}^t$ to replace $\bold{s}$. Results in Table \ref{tab:knowledge_ablation} indicate that the usage of visual relational knowledge can significantly improve detection performance, especially for unseen object pairs.

We further evaluate the generalization ability of \modelname with different visual relation knowledge graphs, and results are presented in Table \ref{tab:KG_ablation}. To explore the effect of the number of nodes on the performance, we build three control-groups namely 0-hop, 1-hop and ALL. In the 0-hop group, we only keep nodes that are in the tested benchmarks. In the 1-hop group, we keep nodes in 0-hop and those connected by at least one edge to nodes in the 0-hop group (\ie one step reachable). For ALL, we simply keep all nodes in the original full graph.
From the results we can observe that the performance first increases with more nodes from 0-hop (150 classes) to 1-hop (17,501 classes) and then declines when all nodes are included. We argue that this is because more noisy information are introduced when all nodes are kept, leading to worse performance.
To study the effect of more types of relationships, we compared using all relationships with using only the 50 most frequent relationships in Visual Genome.  The results show that more diverse relationships are helpful to improve the VRD performance.

Furthermore, we demonstrate the broad applicability of visual relation knowledge by improving the performance of MotifNet\cite{zellers2018motif} and IMP \cite{xu2017IMP} with it. The results in table \ref{tab:vrk} show that both model can achieve better performance when incorporating extra visual relation knowledge (VRK).

\begin{table}[]
    \centering
    \begin{tabular}{lc | rr | cc}
    \toprule
         Node & FR. & $|\mathcal{V}|$ & $|\mathcal{R}|$ & psR & puR \\
    \midrule
         0-hop   & \checkmark &    150 & 35  & 61.3 & 36.1\\
         1-hop   & \checkmark & 17,501 & 35  & 62.3 & 37.7\\
         ALL     & \checkmark & 30,473 & 35  & 60.8 & 36.1\\
    \midrule
         0-hop   &   &    150 & 1,803 & 61.4 & 37.2 \\
         1-hop   &   & 21,363 & 3,509 & 59.4 & 37.9 \\
         ALL     &   & 36,337 & 4,349 & 62.7 & 37.0\\
    \bottomrule
    \end{tabular}
    \vspace{0.3em}
    \caption{Visual relation knowledge graph ablation study. FR.: Filter relationships.}
    \label{tab:KG_ablation}
\end{table}

\begin{table}[]
    \centering
    \begin{tabular}{l  | cc}
    \toprule
        Model &  psR & puR \\
    \midrule
        MotifNet \cite{zellers2018motif} (+ VRK) & 34.6 (+4.0) & 15.0 (+5.7)\\
    \midrule
        IMP \cite{xu2017IMP} \hspace{7mm}(+ VRK) & 47.0 (+8.2)& 24.0 (+6.0)\\
    \bottomrule
    \end{tabular}
    \vspace{1.5mm}
    \caption{Visual relation knowledge applicability study.}
    \label{tab:vrk}
\end{table}

\subsection{Qualitative Results}

Figure \ref{fig:case} illustrates some predictions generated by \modelname and the baseline \cite{xu2017IMP}.
\modelname accurately predicts the \texttt{eating} relationship between pairs such as \texttt{cat-food} and \texttt{giraffe-tree}, despite no object class in these pairs present in the training example of \texttt{eating}.
Moreover, in the third example, the baseline incorrectly predicts the \texttt{lying on} relationship, likely because most training examples of \texttt{lying on} involve \texttt{cat} as the \texttt{Subject}.
However, our framework, augmented with both textual knowledge and visual relation knowledge, correctly predicts the true relationship, showing the effectiveness of our approach.

\section{Related Works}
\noindent\textbf{Visual Relation Detection.} Visual relation detection has been a challenging task due to the long-tail distribution of relationships and the complex relationships between objects.
Early methods \cite{lu2016visual, li2017scene, li2017vip} only use the visual features and word vectors, whose semantics are often not comprehensive enough.
To address this issue, later approaches introduce external knowledge to enhance the performance \cite{dai2017detecting, cui2018context, chen2019knowledge}. Cui \etal \cite{cui2018context} construct a semantic graph with word embedding of object class tags to explore the correlations of classes. Peyre \etal \cite{peyre2019detecting} propose to transfer knowledge learnt from seen objects and relationships to their unseen combinations with an analogy strategy. Recently, some methods are proposed to tackle the long-tail problem to detect more relationships. Dornadula \etal \cite{dornadula2019visual} pre-train the model with samples of frequent relationships.
Li \etal \cite{li2022zero} propose to leverage the lexical knowledge of object tags to reduce the intra-relationship variation. Some other works \cite{chen2019scene, yao2021visual} tackled this problem with large-scale weak supervision.
Though they have achieved promising progress, their performance is constrained by the amount and quality of external data, and they rely on a large amount of resource for training since the supervision from pseudo-labels is noisy. Comparing to \cite{dornadula2019visual, li2022zero, chen2019scene, yao2021visual}, our method combines textual knowledge and visual relation knowledge to enhance the understanding which is effective and easy to scale.

\noindent\textbf{Few-shot Learning.}
Learning models to generalize from a few examples is an important direction. Common supervised learning approaches fail with limited training data. Few-shot learning methods often tackle this problem by combining available supervised information with some prior knowledge \cite{wang2020fewshotsurvey,chen2021zero}. Metric learning and meta-learning are the most important technologies used in few-shot learning. Methods of the former usually measure the similarity between the few-shot examples and the test sample \cite{koch2015siamese, vinyals2016matching, snell2017prototypical}, while methods of the latter aim to learn the optimization of model parameters \cite{finn2017model, ravi2017optimization}.  Our framework more closely resembles the former because it learns a metric module to pull the input of the same relationship to be close. However, it utilizes richer knowledge from pre-trained language models and a visual relation knowledge graph.


\section{Conclusion}
In this work, we propose a novel framework incorporating both textual knowledge and visual relation knowledge to improve the generalization ability of few-shot VRD. Comprehensive experiments conducted on three benchmarks show the effectiveness of our framework. In the future, we plan to acquire knowledge from more forms and explore more advanced knowledge fusion approaches.

{\small
\bibliographystyle{ieee_fullname}
\bibliography{egbib}
}

\end{document}